\documentclass[letterpaper, 10 pt, conference]{ieeeconf}  

\IEEEoverridecommandlockouts                              

\usepackage{color}
\usepackage{amsmath} 
\usepackage{graphicx}
\usepackage{subcaption}  

\title{\LARGE \bf
Detection-Tracking for Efficient Person Analysis: The DetTA Pipeline
}

\author{Stefan Breuers$^{1}$, Lucas Beyer$^{1}$, Umer Rafi$^{1}$, Bastian Leibe$^{1}$
\thanks{Both first authors contributed equally to this work, which has been funded, in parts, by ARA1 BMBF project FRAME (16SV7840) and ERC Consolidator Grant project DeeViSe (ERC-2017-COG-773161).}
\thanks{$^{1}$RWTH Aachen University, Visual Computing Institute
        {\tt\small \{lastname\}@vision.rwth-aachen.de}}%
}

\begin{document}

\maketitle
\thispagestyle{empty}
\pagestyle{empty}

\begin{abstract}
In the past decade many robots were deployed in the wild, and people detection and tracking is an important component of such deployments.
On top of that, one often needs to run modules which analyze persons and extract higher level attributes such as age and gender, or dynamic information like gaze and pose.
The latter ones are especially necessary for building a reactive, social robot-person interaction.

In this paper, we combine those components in a fully modular detection-tracking-analysis pipeline, called DetTA.
We investigate the benefits of such an integration on the example of head and skeleton pose, by using the consistent track ID for a temporal filtering of the analysis modules' observations, showing a slight improvement in a challenging real-world scenario.
We also study the potential of a so-called ``free-flight'' mode, where the analysis of a person attribute only relies on the filter's predictions for certain frames.
Here, our study shows that this boosts the runtime dramatically, while the prediction quality remains stable.
This insight is especially important for reducing power consumption and sharing precious \mbox{(GPU-)memory} when running many analysis components on a mobile platform, especially so in the era of expensive deep learning methods.


\end{abstract}

\section{INTRODUCTION}
\label{sec:intro}

Full detection-tracking systems are a key component in modern robotics.
Being aware of all present persons is important for both navigation as well as interaction of a mobile robot platform.
Especially human-robot interaction can be more responsive and sophisticated if person attributes such as age, gender, appearance, orientation or pose are observed.
As noted by Sabanovic~\emph{et~al.}~\cite{sabanovic2006robots} design choices such as utilizing ``gaze as a sign of interest in interaction [may] improve [the] robot's interactive effectness'', helping person robot-interaction scenarios like person-following~\cite{alvarez2012feature,gockley2007natural} or approaching~\cite{dautenhahn2006may,ramirez2016robots}.

In the area of computer vision, these person analysis modules are usually evaluated on a frame-by-frame basis on controlled test setups.
Running them this way on robots that interact with multiple people at once, person ID and temporal information would be ignored.
With deep learning methods on the rise, these components also introduce a high power consumption and (GPU-)memory usage, making it difficult to run them on low-power mobile platforms with a focus on long-term autonomy~\cite{hawes2016strands}.

In this work, we address those problems by presenting a fully modular detection-tracking-analysis pipeline (DetTA).
Our contributions are therefore i) a fully integrated detection-tracking-analysis pipeline as real-time ROS component, ready to run on a robot, ii) studying the effectiveness of coupling person attribute observations with a track-based temporal filter on the example of head orientation and skeleton pose, and iii) enabling a so-called ``free-flight'' mode, which runs the analysis modules with a stride and relying on the filter's predicition in-between, leading to a dramatic performance boost while keeping stable quality.


After discussing the related work in Sec.~\ref{sec:relwork}, we present our DetTA pipeline in Sec.~\ref{sec:detta_pipeline}.
It is then evaluated in Sec.~\ref{sec:exp_res}, especially looking at the temporal filtering and the ``free-flight'' option.
Sec.~\ref{sec:disc} discusses applications in (social) robotics and future work, before Sec.~\ref{sec:conc} concludes the paper.

\begin{figure}[t]
\centering
\includegraphics[width=0.95\columnwidth]{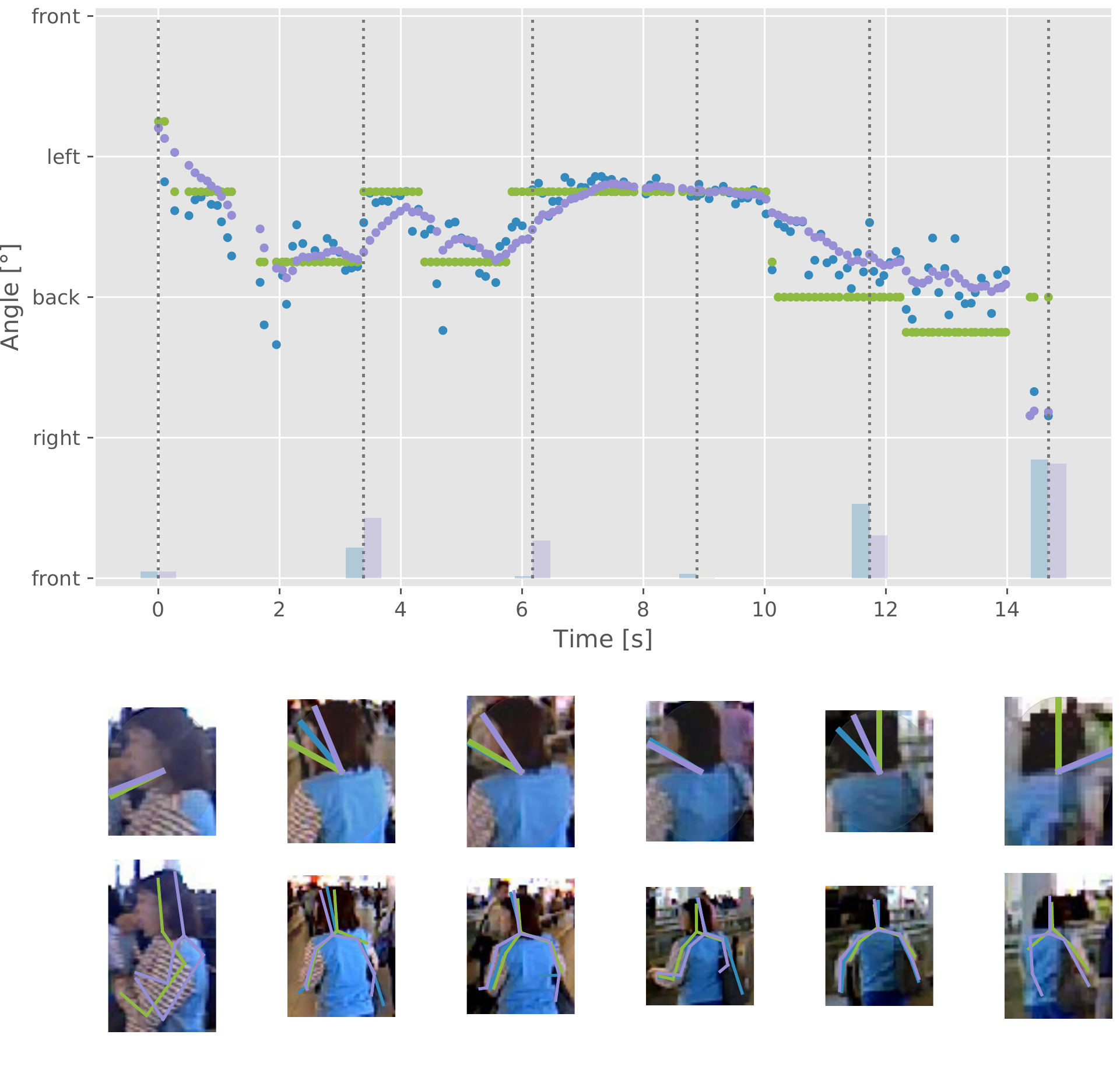}
\vspace{-0.4cm}\caption{%
    Head orientation and skeleton estimated on top of a person's track.
    Green corresponds to the ground-truth annotation, blue to the analysis module's output, and purple to the track-smoothed value.
    The dotted vertical lines show where in time the pictures on the bottom are located, and the bars around these correspond to raw and smoothed angular error, respectively.
    Best viewed on a screen.}
\label{fig:covergirl}
\end{figure}

\begin{figure*}[t]
\centering
\includegraphics[width=0.95\textwidth]{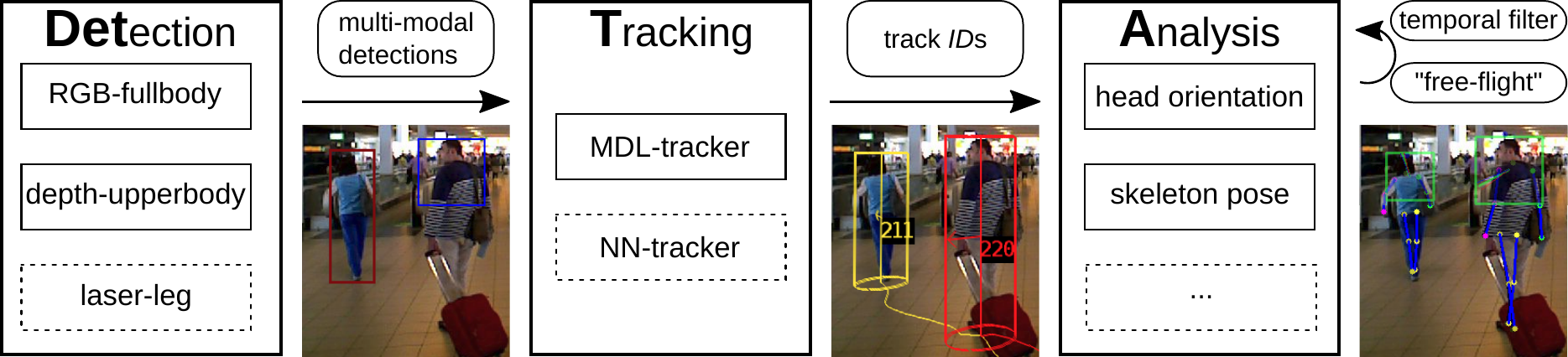}
\caption{%
    The presented pipeline.
    The analysis modules build upon an established detection-tracking framework.
    This allows for both cropped image boxes serving as input, while the consistent track ID can be used for temporal integration of person attributes.
    Dotted components are not considered in this paper, but are part of the pipeline and can readily be used, e.g., to integrate more analysis modules.}
\label{fig:pipeline}
\end{figure*}

\section{RELATED WORK}
\label{sec:relwork}

People awareness is an essential component of any social robot deployed in the wild, as done in recent research projects such as EUROPA~\cite{kummerle2013navigation}, STRANDS~\cite{hawes2016strands} or SPENCER~\cite{triebel2016spencer}.

Full detection-tracking pipelines for robotic scenarios have recently been described and evaluated in~\cite{lindermulti}, including two nearest-neighbor methods~\cite{linder2015towards,dondrup2015real}, the Multi-Hypothesis Tracker of Arras~\emph{et~al.}~\cite{arras2008efficient} and a vision-based MDL-tracker~\cite{jafari2014real}.
In the area of computer vision, the MOTChallenge~\cite{MOTChallenge2015,MOTChallenge2016} gives a good overview of recent image-based tracking methods, but many of those are not designed for robotics purposes as they are not capable of running online.


Person analysis is a broad field of research, as so many different attributes can be of interest depending on the scenario.
Reaching from static information such as gender~\cite{linder2015real}, identity~\cite{HermansBeyer17Arxiv}, clothing or hairstyle~\cite{linder2015real2}, to dynamic information such as gaze direction~\cite{beyer2015biternion,benfold2011unsupervised,benfold2009guiding} or pose skeletons~\cite{rafi2016efficient,newell2016stacked,wei2016cpm}.
But these works only evaluate their method frame-by-frame for a fixed test benchmark, whereas we want to study how those analysis modules behave when run continuously, also enabling temporal smoothing.

To the best of our knowledge, there are not many works which take a full system-level look at how temporal integration can help analysis modules inside a detection-tracking pipeline in a real-world robotic scenario.
Song~\emph{et~al.}~\cite{song2011multi} looks at the specific problem of gesture recognition, which is ``significantly improved by Gaussian temporal smoothing,'' but test this in a very controlled environment with only a single person in the center.
Furthermore, Wall~\emph{et~al.}~\cite{wall2017online} takes temporal integration into account for nod detection in a face-to-face robot-person interaction.

Another related approach by Benfold~\emph{et~al.}~\cite{benfold2011unsupervised,benfold2009guiding} combines a tracking system with gaze estimation.
They strongly couple the gaze estimation and tracker using hand-engineered assumptions in order to learn a scene-specific gaze estimator for static-camera surveillance scenarios.
This is also the case for the work of Kleinehagenbrock, Fritsch, \emph{et~al.}~\cite{kleinehagenbrock2002person,fritsch2003multi},
where they use spatio-temporal integration to combine a face and a laser-leg detector, to track a person and report the rotation, distance and height of the face.

In contrast to the above, our pipeline allows for modular person analysis components, while we study how those can profit from temporal filtering using the tracker's consistent ID.


\section{DETECTION-TRACKING-ANALYSIS PIPELINE}
\label{sec:detta_pipeline}

\subsection{Overview}
Fig.~\ref{fig:pipeline} gives on overview of our detection-tracking-analysis pipeline ``DetTA''.
The detection component gives candidate locations of people present in the scene, while the tracker connects detections over time, bridging missing detections, identifying false alarms, and assigning a unique track ID to each individual person.
By this, the tracker can provide cropped regions in form of bounding boxes, which serve as input for further image-based per-person analysis modules.
Using the track ID, the information of different person attributes can be filtered over time or used in a ``free-flight'' mode, relying on the filter's predictions.
The pipeline is highly modular and can be extended with different detectors, trackers, and even more analysis modules.
The code repository, which includes ROS nodes ready to run on a robot, will be made publicly available\footnote{https://github.com/sbreuers/detta.git}.

In the following, we describe the integrated methods of this paper, which have been developed during the EU projects STRANDS~\cite{hawes2016strands} and SPENCER~\cite{triebel2016spencer} and were deployed multiple times in an office space, an elderly care home, and a busy airport environment.


\subsection{Detection}
Analysis modules typically extract person attributes from RGB(-D) images.
This is why we use two vision-based person detectors to extract 2D bounding boxes of potential people in the scene, serving as the input for vision-based trackers~\cite{breuers2016exploring, MOTChallenge2015, MOTChallenge2016}.
The first one is based on depth templates of upper bodies~\cite{jafariICRA14}, focusing on detecting persons close to the robot (blue rectangle in the left-most image in Fig.~\ref{fig:pipeline}).
The second one is the groundHOG implementation of~\cite{sudoweCVS2011efficient}, which relies on HOG-features~\cite{dalal2005histograms} and takes advantage of geometrical constraints to detect persons in RGB further away from the robot (red rectangle in Fig.~\ref{fig:pipeline}).
Both detection channels serve as input for the tracking module, and by including ground-plane information, 3D world-coordinates of tracks are available.

\subsection{Tracking}
As a tracking method, we rely on the MDL-tracker described in~\cite{jafariICRA14}, which has shown competitive performance in real-world scenarios~\cite{lindermulti}.
It is based on the approach of Leibe~\emph{et~al.}~\cite{leibePAMI08}, following the framework of~\cite{essPAMI2009robust,schindlerISPRS2010automatic} to compute an overcomplete set of multiple track hypotheses, similar to the MHT approach~\cite{arras2008efficient}.
New trajectories are added to this set by a bi-directional Extended Kalman Filter following a constant velocity motion model backwards in time.
Existing trajectories are extended from the last to the current frame, also following this motion model.
Each track in the overcomplete set gets a score based on confidence, appearance, and motion agreement of inlying detections, while physical overlap and shared detections define an interaction cost between trajectories.
The problem of choosing the best subset is then formulated as quadratic binary problem solved by the multi-branch method of~\cite{schindlerECCV2006perspective}.

It should be mentioned that our tracking pipeline is multi-modal and allows for the inclusion of, e.g., laser-based leg-detection~\cite{Beyer16RAL,Arras07ICRA,Leigh15ICRA} in addition to the aforementioned vision-based detectors.
This can help the tracker preserve track IDs when the persons leave the camera's field-of-view, and can additionally be used for analysis modules which do not require image data.

\subsection{Analysis}
While it is not a restriction of our framework, we limit the analysis modules to head orientation and upper-body pose skeletons for this work.

We predict head orientation using BiternionNets~\cite{beyer2015biternion}, for which code is publicly available.
Training data is collected at an airport by having volunteers turn in circles in front of our robot, the annotation is straightforward and done in just a few hours.
Biternions have the advantage of providing continuous head pose estimates, which are better suited for filtering than classes, even when trained on discrete labels.
The network architecture is exactly the very lightweight one introduced in~\cite{beyer2015biternion}, but we further perform background-subtraction using the depth data provided by the camera.


For skeleton poses, we use the HumanPose estimation framework from \cite{rafi2016efficient}.
The framework is an adaptation of GoogleNet \cite{szegedy2015going}, using only the  first 17 layers from the network architecture.
The fully connected layer and the average pooling layer in the last stages of the network are removed to make the framework fully convolutional.
A pose decoder consisting of a transposed convolution and a sigmoid layer is appended to the framework to up-sample the low resolution features from the 17th layer to high resolution heat maps for different body joints.
The HumanPose estimation framework was trained on the MPI dataset \cite{andriluka20142d, pishchulin2014fine} and is also able to detect occluded joints.


 
On top of these image-based analysis modules, further trajectory information, like speed, accelaration, orientation, distance to robot, etc. could be reported.
But as these often can be derived from the tracker's output itself, which ususally includes a temporal filter on its own, it is not part of our experiments.
We want to couple the track ID with the output of indiviual modular analysis components to perform the temporal integration.

\subsection{Temporal Filtering}
As mentioned, analysis methods are typically designed and evaluated on an individual, frame-by-frame basis.
Here, the tracking step of the DetTA pipeline provides consistent person IDs which enables keeping the observed information and integrating it over time, as long as this ID persists.
So for each ID currently present and for each analyzed person attribute, we generate an individual filter.
Many more details on filtering methods than discussed here can be found in~\cite{labbe2015kalman}.

The most straightforward approach when it comes to filtering a single observed quantity is the so called g-h-filter, also known as $\alpha$-$\beta$-filter or a-b-filter~\cite{penoyer1993alpha}.
The state variable is modeled as $(x, v)$, i.e., the filtered value and its first derivative/velocity.
Two parameters, $g$ and $h$, represent the update rate of the value and its derivative, respectively, leading to the following predict-update loop given observation $z$:

\begin{align}
    \tilde{x}_t &= x_{t-1} + v_{t-1} \Delta t\\
    x_t &= \tilde{x}_t + g (z_t - \tilde{x}_t)\\
    v_t &= v_{t-1} + h \frac{(z_t - \tilde{x}_t)}{\Delta t}
\end{align}

It is similar to a weighted average between the predicted next state and the incoming measurement.
Extension to higher-order derivatives is straight-forward but typically unstable due to the estimation of high-order derivatives from noisy data.

While our code framework also includes more sophisticated filtering methods, like the well-known Kalman Filter~\cite{kalman1960new,welch2006introduction}, we observed in our explorative study that the g-h-filter is already able to capture the occuring motion behaviour for our used analysis modules.
Of course, this does not hold for all attributes and needs to be investigated on a case-by-case basis.
The Kalman Filter can actually be derived from the g-h-filter by updating $g$ and $h$ on-the-fly, allowing for a richer state representation, motion models, and adaptive correlations at the cost of worse generalization due to more hyperparamters which need to be tuned.
Furthermore, its assumption of uncorrelated Gaussian noise is violated by most existing analysis modules, and it can be difficult to find a motion model accurately reflecting the real-world behaviour of, \emph{e.g.}, a person's wrist.

As an additional option in our pipeline, we propose the so called ``free-flight'' mode.
Here, the analysis module is only run with a certain \textit{stride}, thus updating the filter with an observation less frequently.
Inbetween those updates, the analysis information relies on the filter's predictions.
This allows for severe computational savings, increasing performance, reducing power-consumption and (GPU-)memory usage, especially for expensive analysis components.

\begin{figure}
	\centering
	\begin{subfigure}[b]{0.24\textwidth}
		\centering
		\includegraphics[height=2.5cm]{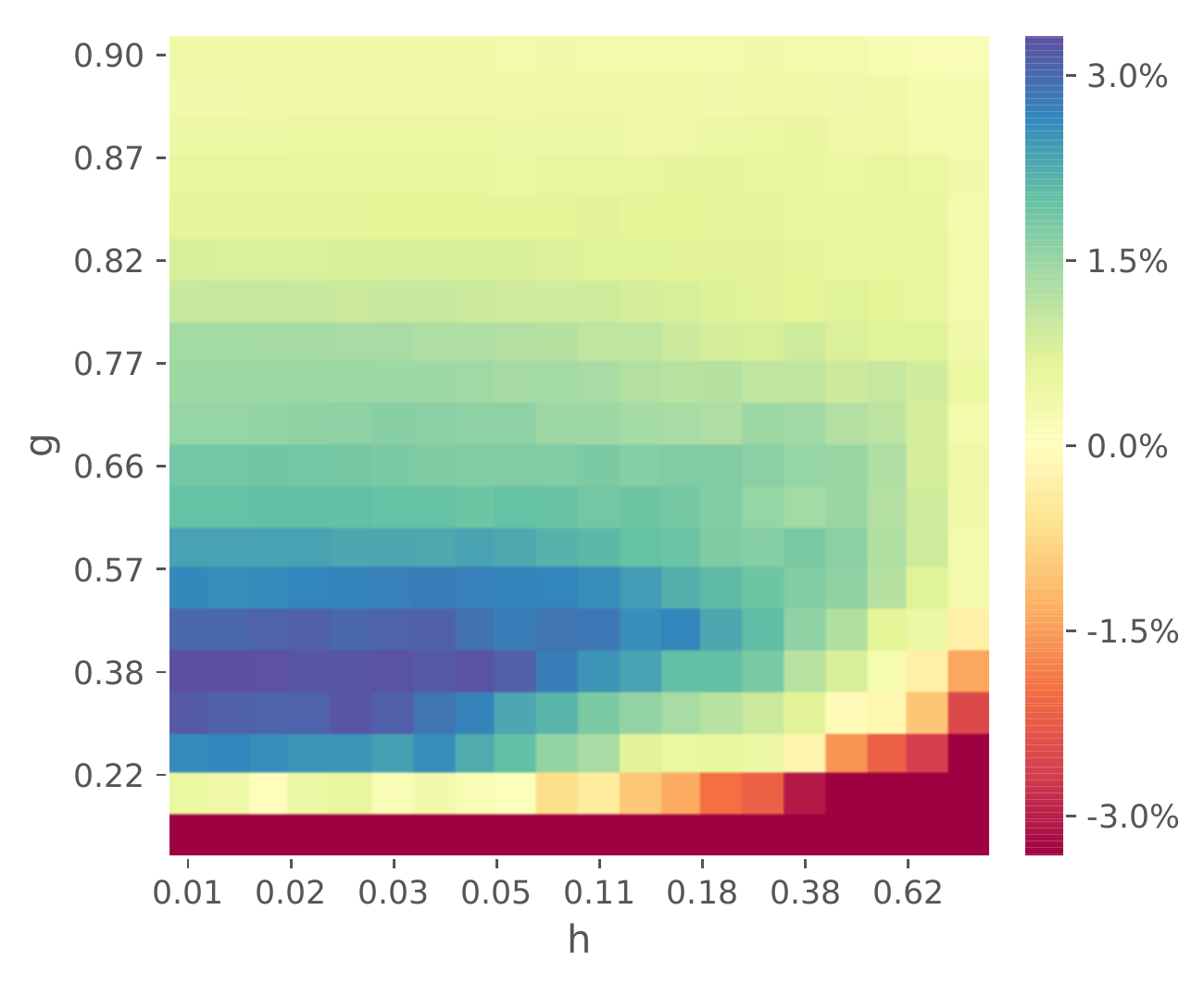}
		\caption{}
		\label{fig:heads_gh_heatmap}
	\end{subfigure}%
	\begin{subfigure}[b]{0.24\textwidth}
		\centering
        \begin{tabular}{c|cc}
        & offset & PCO\\\hline
        raw & $38.9^{\circ}$  & 64.9\%\\
        filter & $36.8^{\circ}$ & 67.7\%
        \end{tabular}
		\caption{}
		\label{fig:heads_gh_table}
	\end{subfigure}
    \caption{Filtering results for heads, exploring the impact of different $g$, $h$ values for head orientation quality in (a) and numeric results for the chosen values of $g=0.5$ and $h=0.02$ in (b).
    Colors on the blue side of the spectrum represent improvement over raw predictions, while colors on the red side represent worse results. A $g$ value of $0.0$ is a constant prediction.}
	\label{fig:heads_filt_res}
\end{figure}

\section{EXPERIMENTS AND RESULTS}
\label{sec:exp_res}

\subsection{Setup}
For the quantitative evaluation, we have chosen a real-world scenario recorded in a busy airport environment as part of the SPENCER project~\cite{triebel2016spencer}.
People are getting off a flight, stepping in front of the camera, looking around and pointing directions, which is well suited for the evaluation of our used analysis modules.
The scene was recorded with an Asus Xtion Pro Live RGB-D sensor mounted on a mobile platform.
We selected a subsequence of 1218 frames and, for a total of 77 persons, carefully annotated their 2D bounding box tracks using AnnoTool2~\cite{MilanSchindlerRothCVPR13}, their head orientations as described in~\cite{beyer2015biternion}, and their skeletons poses with a custom annotation tool, which is part of our code repository.
For the skeleton, we only annotated the 8 upperbody joints (head, neck, left and right shoulders/elbows/wrists), as these are the most interesting for interaction and the leg joints are not visible for people close to the camera.

\begin{table}[b]
\centering
\begin{tabular}{c|cc|cc|c|c}
\textbf{GT tracks} & \textbf{MOTA} & \textbf{MOTP} & \textbf{FP} & \textbf{FN} & \textbf{IDS (rate)} & \textbf{Hz} \\\hline
77 & 36.4\% & 73.5\% & 616 & 4675 & 120 (1.4\%) & 96
\end{tabular}
\caption{Tracking results for the test sequence.}
\label{tab:track_results}
\end{table}

This ground truth (GT) allows for a reasonable evaluation of the individual analysis components.
For skeletons, we use the widely used PCKh measure~\cite{andriluka20142d}, corresponding to the percentage of correct keypoints, with a threshold normalized by 50\% of the head size (higher is better).
For the head orientation, we define a similar measure: ``PCO,'' the percentage of correct orientations with a threshold of $45^{\circ}$, or $1/8$ of a full circle (higher is better).
We also report the positional or angular offset, respectively, to be able to highlight even slight improvements in the precision (lower is better).

While we discuss the experimental result of the analysis modules in detail in the following sections, we briefly mention the tracking performance first, as the analysis modules build up on that.
We follow the established CLEAR metrics~\cite{BS2008}, computing the number of false positives (FP), false negatives (FN) and ID switches (IDS) which are then summarized by the tracking accuracy MOTA and precision MOTP.
Tab.~\ref{tab:track_results} shows the results averaged over 5 runs to account for synchronization issues in ROS.
An interesting quantity in our case is the number of IDS, as a new filter is started as soon as an ID switch happens, interrupting the current temporal integration of person attribute information.
As we can see, for 77 ground truth tracks we get 120 ID switches in total, corresponding to 1.6 ID switches per person on average over all 1218 frames.
This is a bearable amount, especially if those switches happen at the beginning or end of the track.

\subsection{Robust Person Analysis}
In this section, we evaluate the analysis components, namely head orientation and skeleton pose estimation, and investigate how the filtering of those informations can increase their quality.

We choose $g=0.5$, which equally trust measurements and predictions, while an $h$ of $0.02$ accounts for slight angular velocities when a persons looks around.
While these values seem reasonable, Fig.~\ref{fig:heads_gh_heatmap} which explores different $g$ and $h$ values in an offline experiment, shows that the method is not too sensitive to their exact choice.

For skeleton poses, different joints follow different motion models.
Head, neck and elbows have a low to none ego-motion and mostly follow the person's constant velocity itself.
This is why we set $h$ to a somewhat higher value of $0.2$.
With a $g$ of $0.9$, we lay more trust on our analysis method, also to re-align occluded joints.
Wrists have the strongest ego-motion, so we set a lower $h=0.02$ and $g=0.5$.
Again, we show an exploration of the $g$-$h$-space on the representative example of the wrist in Fig.~\ref{fig:heads_gh_heatmap}.


\begin{figure}
	\centering
	\begin{subfigure}[b]{0.24\textwidth}
		\centering
		\includegraphics[height=2.5cm]{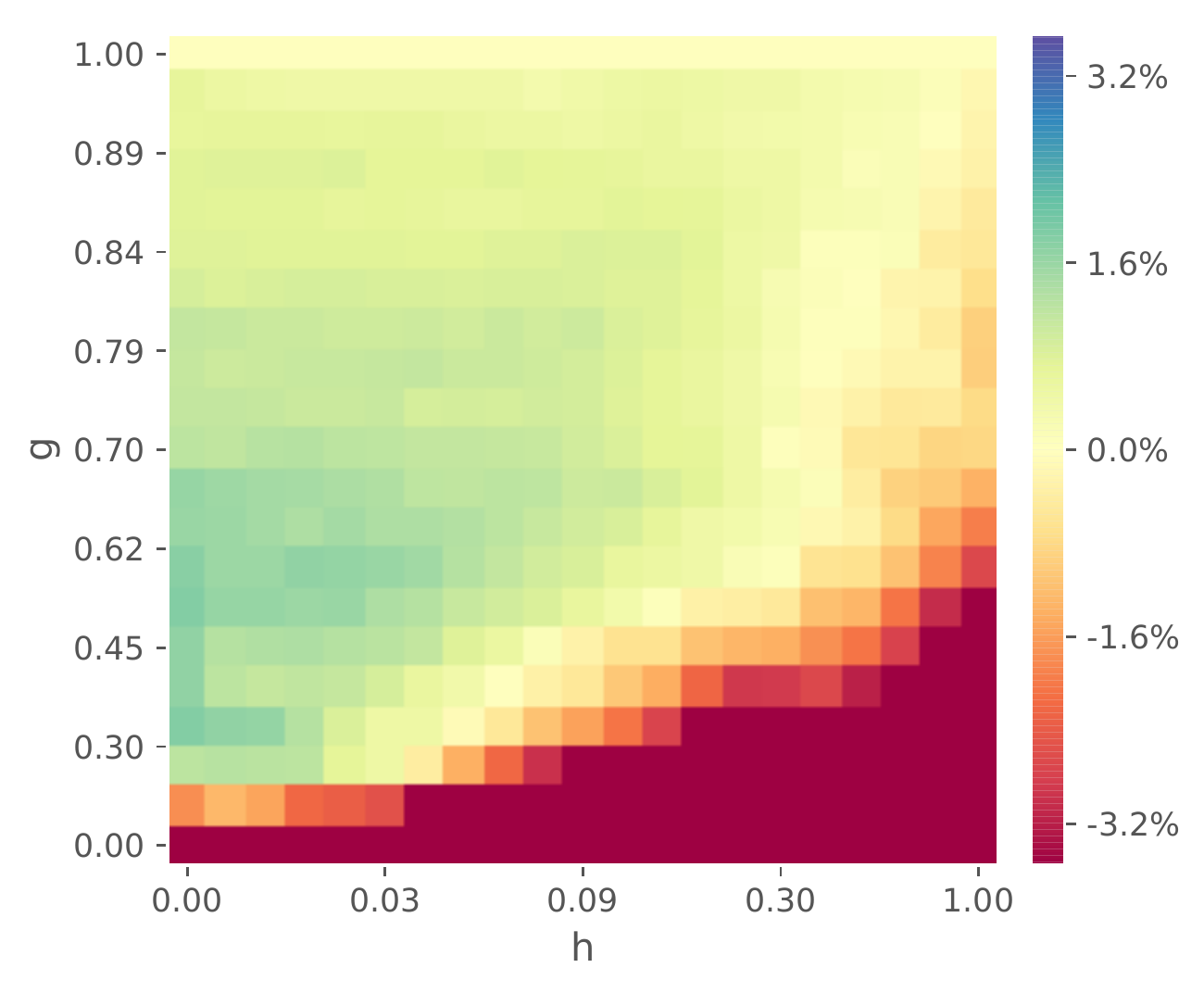}
		\caption{}
		\label{fig:skels_gh_heatmap}
	\end{subfigure}%
	\begin{subfigure}[b]{0.24\textwidth}
		\centering
        \begin{tabular}{c|cc}
        & offset & PCKh\\\hline
        raw & 13.1 px  & 72.1\%\\
        filter & 12.6 px & 73.6\%
        \end{tabular}
		\caption{}
		\label{fig:skels_gh_table}
	\end{subfigure}
    \caption{The same analysis as in Fig.~\ref{fig:heads_filt_res} done on the representative example of wrist in (a) and numeric results for the chosen values of $g=0.5$ and $h=0.02$ in (b).}
	\label{fig:skels_filt_res}
\end{figure}

Tab.~\ref{fig:heads_gh_table} or Tab.~\ref{fig:skels_gh_table}, respectively, report the performances for our chosen values.
We see that the values indeed slightly increase.
Regarding the rather minor quality gain, we want to remark that for perfect analysis modules, a filter cannot improve the result, which also holds for bad performing observations with systematic noise.
Also, the absolute numbers depend on the strictness of the used metrics.


Still, our method holds the potential for a performance increase, which will be discussed in the next chapter.

\begin{table*}
\centering
\begin{tabular}{c|cc|cc|cc|cc|ccc}
 &  \multicolumn{2}{c|}{\textbf{Head/Neck (PCKh)}} & \multicolumn{2}{c|}{\textbf{Shoulders (PCKh)}} & \multicolumn{2}{c|}{\textbf{Elbows (PCKh)}} & \multicolumn{2}{c|}{\textbf{Wrists (PCKh)}} & \multicolumn{3}{c}{\textbf{Performance (Hz)}} \\\hline
 stride & keep & predict & keep & predict & keep & predict & keep & predict & GTX 1080Ti & Mobile GT 730 & i7 2.4GHz$\times$8 \\\hline
 1 & 74.8 & 74.8 & 74.6 & 74.6 & 69.4 & 69.6 & 72.1 & 73.6 & 111 & 108 & 256\\
 2 & 64.3 & 68.8 & 70.0 & 71.2 & 69.6 & 65.5 & 71.8 & 73.2 & 432 & 355 & 423\\
 3 & 55.9 & 62.3 & 64.5 & 65.7 & 61.4 & 62.1 & 70.2 & 71.4 & 532 & 487 & 478\\
 5 & 44.4 & 51.2 & 58.4 & 57.4 & 55.4 & 54.0 & 68.4 & 67.9 & 584 & 574 & 583\\
\end{tabular}
\caption{Analysis of free-flight mode for skeleton joints}
\label{tab:skels_ff_results}
\end{table*}

\subsection{Free flight}
We now turn to evaluating the ``free-flight'' option described at the end of Sec.~\ref{sec:detta_pipeline}.
To paint a complete picture, we look at the performance using three different hardware setups: a high-end desktop GPU like the NVIDIA~GTX~1080Ti, a low-power mobile GT 730M GPU, and a high-end but mobile Intel i7 CPU.
Note that the full pipeline, including CUDA-based detection, is running during those experiment, and hence the system is under load even without the analysis components.

Tab.~\ref{tab:heads_ff_results} shows the full comparison for our head orientation analysis component.
As we can see, the ``free-flight'' mode with a stride of 2 already boosts the runtime by a factor of around 5 for the GPUs and 10 for the CPU.
When increasing the stride, we reach a point where we hit a ceiling in the performance, while the quality starts dropping (64.8 vs. 62.0).
Overall, including a predictive motion model (``predict'' column) always increases the PCO measure during free-flight as compared to a naive baseline of only sticking to the last observed measurement (``keep'' column), which corresponds to a g-h-filter with $g=1$ and $h=0$.

For skeleton poses, Tab.~\ref{tab:skels_ff_results} shows similar results.
The overall quality regarding the PCKh measure decreases faster with higher strides, confirming that a simple motion model struggles with describing behaviour as complex as joints' motion.
This even leads to a better quality of the ``keep'' baseline in the case of elbows (stride 2 and 5) and wrists (stride 5).
Still, considering all joints, the quality does not drop critically with a stride of 2, while the performance increases noticeably, up to a factor of 4 for the GTX 1080Ti.

These experiments confirm the effectiveness of the free-flight mode.







\begin{table}
\centering
\begin{tabular}{c|cc|ccc}
 &  \multicolumn{2}{c|}{\textbf{Quality (PCO)}} & \multicolumn{3}{c}{\textbf{Performance (Hz)}} \\\hline
    stride & keep & predict & GTX 1080Ti & GT 730M & i7 2.4GHz$\times$8 \\\hline
 1 & 64.9 & 67.7 & 395 & 325 & 157\\
 2 & 64.5 & 66.9 & 1866 & 1551 & 1531\\
 3 & 63.3 & 64.8 & 2316 & 1940 & 1865\\
 5 & 61.9 & 62.0 & 2108 & 2261 & 2271\\
\end{tabular}
\caption{Analysis of free-flight mode for head pose}
\label{tab:heads_ff_results}
\end{table}

\section{DISCUSSION}
\label{sec:disc}
Our temporal filtering of several person attributes may support any social robot platform, while the free-flight helps saving ressources and battery-life.
Especially mobile platforms that do not come with a high end GPU or are desigend for long-term autonomy can profit from this performance boost.
Our DetTA framework can help improve any social robot-person interaction, like person following and approaching, returning a look, interpreting gestures, etc.
A concrete use case for this is addressed in a very recent project FRAME\footnote{www.frame-projekt.de} this work is partially sponsored by.
Here, the robot has to identify potential helpers in a scene and ask them to open doors and operate elevators.
The robust temporal integration of person analysis information helps in the decision-making process, e.g., to extract awareness information and whether a person shows interest in the robot.

A promising direction for research beyond the application point of view is to gauge how well a measurement matches the current filter state and use this information to compute a distinct confidence score for analysis modules.
This is especially useful for deep learning methods that often struggle with providing a useful confidence.
Certainly, one could add even more analysis models, also considering static attributes (age, gender, etc.) in a majority voting fashion which gathers evidence over time.
Finally, as hinted by the results on skeletons, some attribute ``motion'' can get very complex and could profit from being learned, a direction we did not yet explore.





\section{CONCLUSIONS}
\label{sec:conc}
In this paper we presented a full detection-tracking-analysis pipeline, combining tracking information with the estimation of person attributes, such as head orientations or skeleton poses.
Here, the consistent trajectory ID of each person allows for a temporal integration of the output of the analysis modules, which could otherwise only operate frame-by-frame.

While this temporal filtering only slightly improves quality, it allows for what we call the ``free-flight'' mode, in which the analysis modules are only run every $x$th frame.
Inbetween those, the estimation of the person attributes relies on the filter's prediction.
The more analysis modules are available, the more resources get constrained, especially with the rise of deep learning based methods.
By using this free flight mode and staggering the ticks at which the modules are evaluated, it is possible to run many more such demanding modules than would otherwise be possible, while still being able to use predictions at every point in time.

Our pipeline allows for modular extension with own custom analysis modules to get temporally robust person attributes for improved social robot interaction in different application scenarios.







{\small
\bibliographystyle{IEEEtran}
\bibliography{detta}
}

\end{document}